\documentclass[10pt,twocolumn,letterpaper]{article}

\usepackage[pagenumbers]{wacv} 

\usepackage{booktabs}
\usepackage{multirow}
\usepackage{graphicx}
\usepackage{colortbl}
\usepackage{arydshln}

\usepackage{cancel}   

\usepackage[normalem]{ulem}
\useunder{\uline}{\ul}{}

\definecolor{wacvblue}{rgb}{0.21,0.49,0.74}
\usepackage[pagebackref,breaklinks,colorlinks,allcolors=wacvblue]{hyperref}
\newcommand\blfootnote[1]{%
    \begingroup
    \renewcommand\thefootnote{}\footnote{#1}%
    \addtocounter{footnote}{-1}%
    \endgroup
}

\def\wacvPaperID{2121} 
\def\confName{WACV}
\def\confYear{2026}

\title{EmojiDiff: Advanced Facial Expression Control with High Identity Preservation in Portrait Generation}

\author{
 Liangwei Jiang \quad
 Ruida Li \quad
 Zhifeng Zhang \quad
 Shuo Fang\textsuperscript{\rm \dag} \quad
 Chenguang Ma\textsuperscript{\rm \dag} 
\vspace{1mm}
\\
\fontsize{11pt}{13pt}\selectfont{Terminal Technology Department, Alipay, Ant Group}\\
{\tt\small
\{jiangliangwei.jlw, ruida.lrd, jason.zzf, fangshuo.f, chenguang.mcg\}@antgroup.com}
}

\begin{document}
\maketitle
\blfootnote{{\rm \dag} Corresponding authors.}

\vspace{-0.2cm}
\begin{abstract}
This paper aims to bring fine-grained expression control while maintaining high-fidelity identity in portrait generation. This is challenging due to the mutual interference between expression and identity. On one hand, fine expression control signals inevitably introduce appearance-related semantics (\textit{e.g.}, facial contours, and ratio), which impact the identity of the generated portrait. On the other hand, even coarse-grained expression control can cause facial changes that compromise identity, since they all act on the face. Here, we introduce \textbf{EmojiDiff}, the first end-to-end solution that enables simultaneous control of extremely detailed expression  (RGB-level) and high-fidelity identity in portrait generation. To address the above challenges, EmojiDiff adopts a two-stage scheme involving decoupled training and fine-tuning. For decoupled training, we innovate \textbf{I}D-irrelevant \textbf{D}ata \textbf{I}teration (IDI) to synthesize high-quality cross-identity expression pairs by separating and optimizing the processes of maintaining expression and altering identity. Training the model with this data, we effectively disentangle fine expression features in the expression template from other extraneous information (\textit{e.g.}, identity, skin). Subsequently, we present \textbf{I}D-enhanced \textbf{C}ontrast \textbf{A}lignment (ICA) for further fine-tuning. ICA achieves rapid reconstruction and joint supervision of identity and expression information, thus aligning identity representations of images with and without expression control. Experimental results demonstrate that our method significantly outperforms its counterparts, achieving precise expression control with highly maintained identity, and generalizing well to various diffusion models. Project page: \url{https://emojidiff.github.io/}.

    \end{abstract}

\vspace{-0.2cm}
\begin{figure}[htp]
  \centering
  \includegraphics[width=0.95\columnwidth]{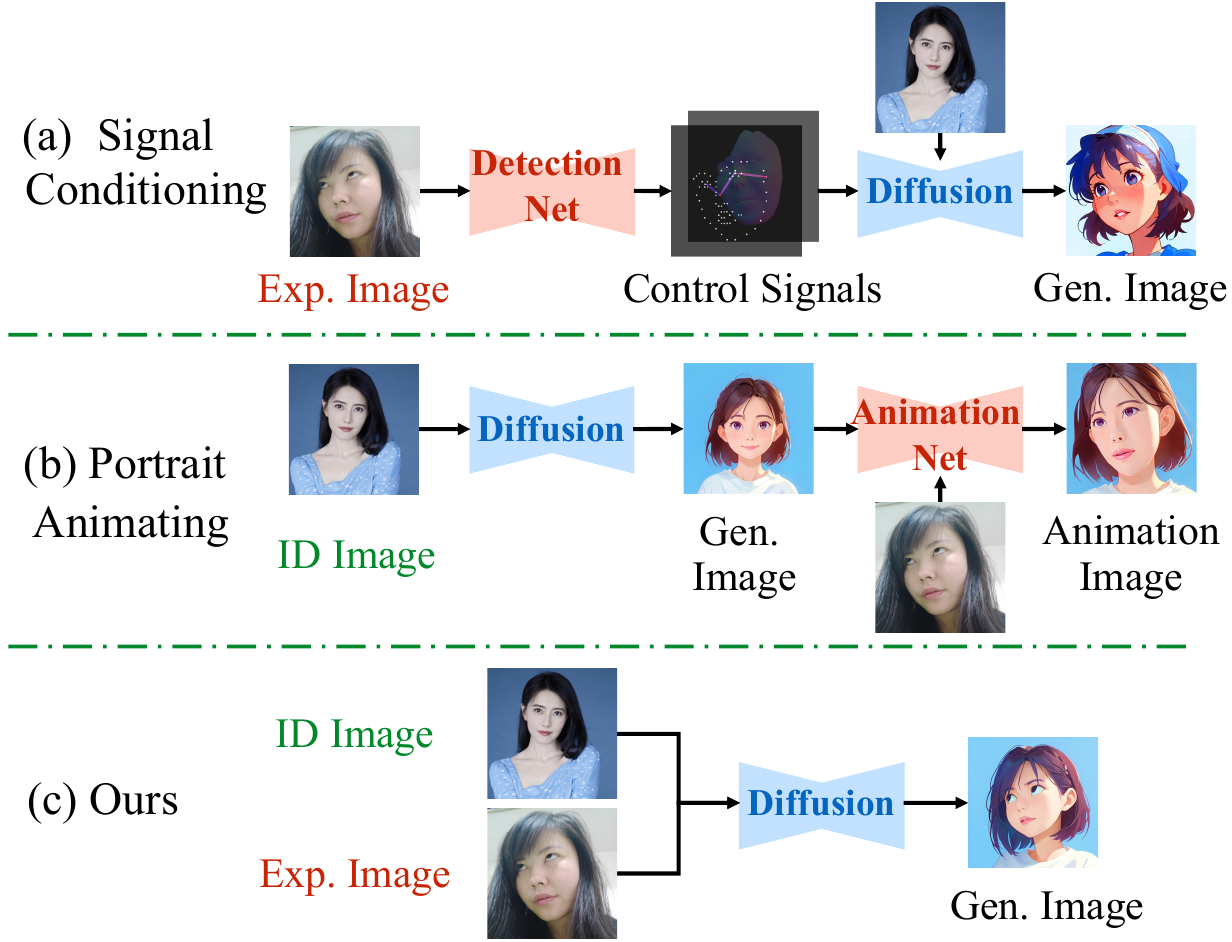}
  \caption{\textbf{Expression customization by different methods}. (a) The methods extract the control signals from expressions and inject them into diffusion models. (b) The methods generate stylized images and manipulate them through animation. (c) The proposed method simultaneously incorporates portrait images and reference expressions, generating stylistic images in an end-to-end manner.} 
  \label{fig:banner}

  \end{figure}

\section{Introduction}
\label{sec:intro}
The development of diffusion model \cite{ho2020denoising, rombach2022sd, ruiz2023dreambooth, hu2021lora,nichol2021glide} has catalyzed significant advancements in identity (ID) customization \cite{ye2023ipa,wang2024instantid,guo2024pulid}. This task involves generating images that align with the specified identity of portrait images, with wide-ranging applications in AI portraits \cite{li2024photomaker,liu2023facechain,huang2024consistentid}, image animation \cite{hu2024animateanyone,wei2024aniportrait}, and virtual try-ons \cite{choi2024vton,chong2024catvton,shen2024imagdressing}. Despite substantial advancement, existing methods reveal a notable limitation:  monotonous facial expressions. This shortcoming constrains the expressiveness of the generated portraits, undermining their ability to convey subtle emotions and severely hindering practical applications.

Researchers \cite{deng2019arcface,huang2023collaborative,paskaleva2024unified} seek to use reference images to manipulate portrait expressions. However, due to the strong coupling between expression and identity, it is challenging to introduce precise expression control while preserving high identity fidelity. First, fine-grained expression control conflicts with identity preservation. Excessively detailed expression templates also bring appearance-related semantic and style information (such as facial contours and skin). If the model directly replicates the appearance and structure from the expression reference, it compromises the identity of generated images (referred to as ID leakage). Second, a potential conflict exists between facial expression and identity control since both manipulate the face. Expression control can modify the facial topology and reduce identity fidelity, while identity control may make expressions more neutral.
Third, although the highly flexible diffusion model and Style LoRA enrich the generated portraits, they also bring challenges in the generalizability of identity and expression control across various stylized scenes.

Pioneering works can hardly deal with these challenges altogether. As depicted in Fig. \ref{fig:banner}, signal-conditioning generation methods \cite{deng2019arcface,huang2023collaborative,paskaleva2024unified} tend to extract and inject coarse expression signals (\eg, landmarks, poses) into the diffusion model to prevent ID leakage. However, these signals lose many facial details (\eg, frowning, pouting, and cheek twitching), and they are significantly limited by the accuracy of third-party networks, such as pose \cite{OpenPose} and action unit detectors \cite{chang2023libreface}.
Portrait animation \cite{hu2024animateanyone,chen2024echomimic} can serve as a post-processing tool for expression customization, which transfers expressions from reference templates to portrait images. However, these methods heavily rely on generated stylized images, where the identity and data distribution of the portraits significantly deviate from the real world, negatively impacting identity fidelity, expression control, and image quality.

In this paper, we introduce EmojiDiff, an innovative end-to-end solution that integrates fine-grained expression control with high-fidelity portrait generation.
Unlike previous signal conditioning-based methods \cite{OpenPose,paraperas2024arc2face,varanka2024fineface}, we still insist that the model should derive the motion directly from the original RGB expressions, ensuring comprehensive expression representation and robust generalization \cite{guo2024liveportrait,xie2024xportrait}. To achieve this, we propose a pluggable expression controller (E-Adapter) alongside its decoupled training strategy. As the core of the training, we innovate ID-irrelevant Data Iteration (IDI) to synthesize the identity-expression decoupling data (\ie, cross-identity expression pairs), which are rarely available in real-world data. IDI creatively divides cross-identity expression transfer into two steps: maintaining expressions and modifying identities. By sequentially optimizing the two steps, we achieve a stable and high-quality data supply. Training the controller with these pairs facilitates the transfer of detailed expressions to the generated output while implicitly filtering out identity information concealed in the expressions, thereby preventing ID leakage.
To address the potential conflict between expression and identity, we develop the ID-enhanced Contrast Alignment (ICA) for fine-tuning.
ICA efficiently computes the identity and expression loss by one-step reconstruction, 
promoting the identity under different expressions as consistent as possible.
This process mitigates the negative impacts of expression control on identity and enhances identity fidelity.
In summary, our contributions are fourfold:

1) To the best of our knowledge, EmojiDiff is the first end-to-end framework that achieves both fine-grained expression control and high ID fidelity portrait generation, accommodating various diffusion models and style LoRA.

2) We innovate ID-irrelevant Data Iteration for decoupled training, effectively disentangling fine expression features from RGB expression images and preventing ID leakage effectively.

3) We propose ID-enhanced Contrast Alignment, an efficient fine-tuning strategy designed to ensure generated portraits maintain stable identity across diverse expressions.

4) We introduce the cross-identity expression dataset CIEP100k, containing various expressions and data with the same expression but different identities, facilitating further research on expression-identity disentanglement.

\begin{figure*}[!ht]
  \centering
  \includegraphics[width=0.88\textwidth]{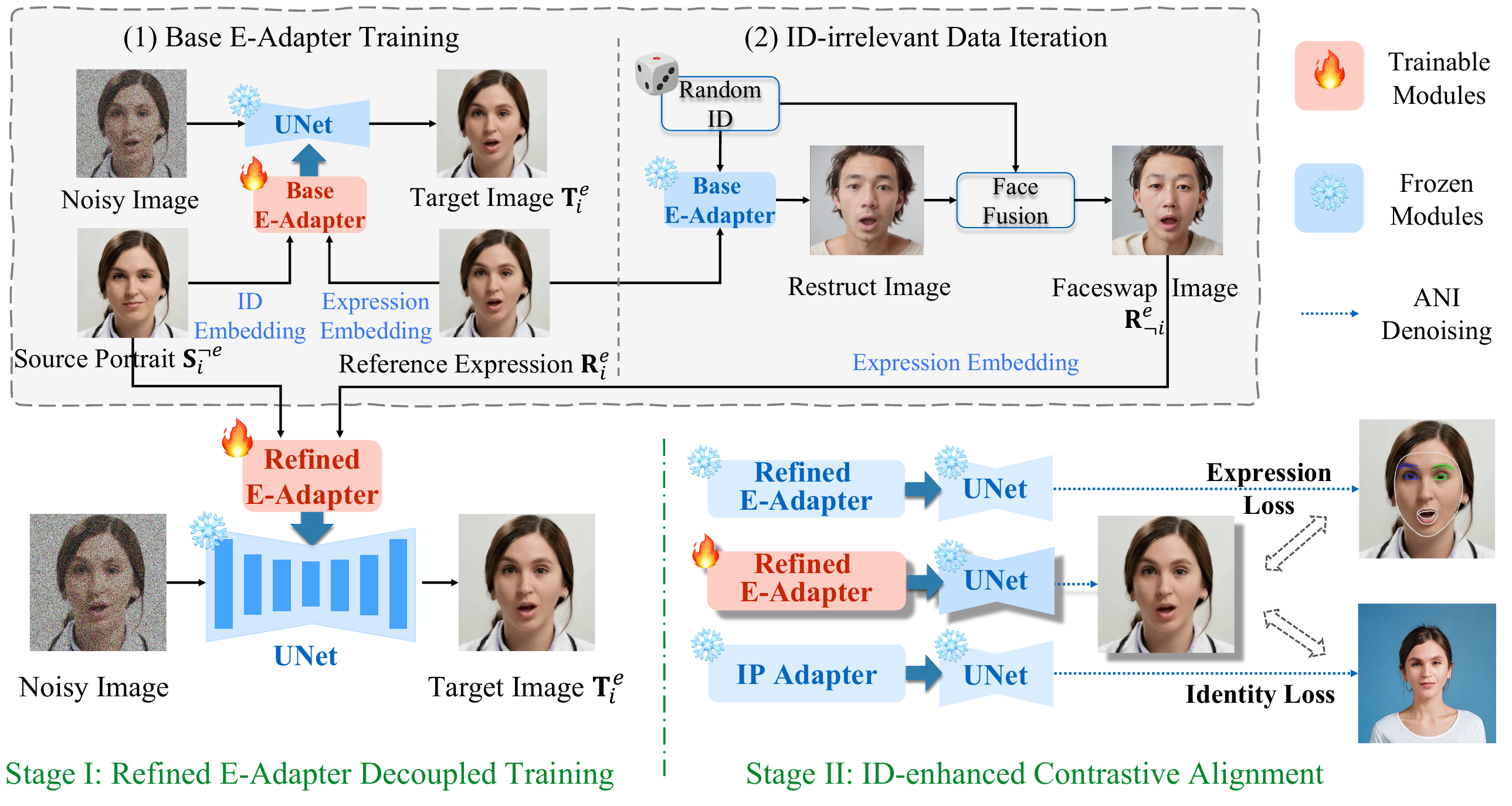}
  \caption{\textbf{Overview of the proposed method}. 
  To integrate RGB-driven expression control into diffusion models, we aim to synthesize cross-identity data $\{\mathbf{S}_i^{\neg{e}}, \mathbf{R}_{\neg{i}}^e, \mathbf{T}_i^e\}$ for the model's decoupled training, and mitigate the negative impact on the original identity through contrastive alignment fine-tuning. Before decoupled training, the fundamental expression controller (\textit{i.e.,} Base E-Adapter) is trained with same-identity data $\{\mathbf{S}_i^{\neg{e}}, \mathbf{R}_{i}^e, \mathbf{T}_i^e\}$ to obtain expression transfer capabilities (the structure of the E-Adapter is illustrated in Fig. \ref{fig:adapter}). Next, the trained Base E-Adapter and FaceFusion \cite{FaceFusion} are utilized to alter the identity of portraits, thereby creating cross-identity expression pairs $\{\mathbf{R}_{\neg{i}}^e, \mathbf{T}_i^e\}$. Subsequently, the Refined E-Adapter uses newly synthesized data for disentangled training, facilitating dual control of identity and expression without ID leakage. Finally, the Refined E-Adapter is fine-tuned by expression and identity loss based on ANI.
 } 
  \label{fig:pipeline}
  \end{figure*}
  \begin{figure}[h]
    \centering
    \includegraphics[width=1.0\columnwidth]{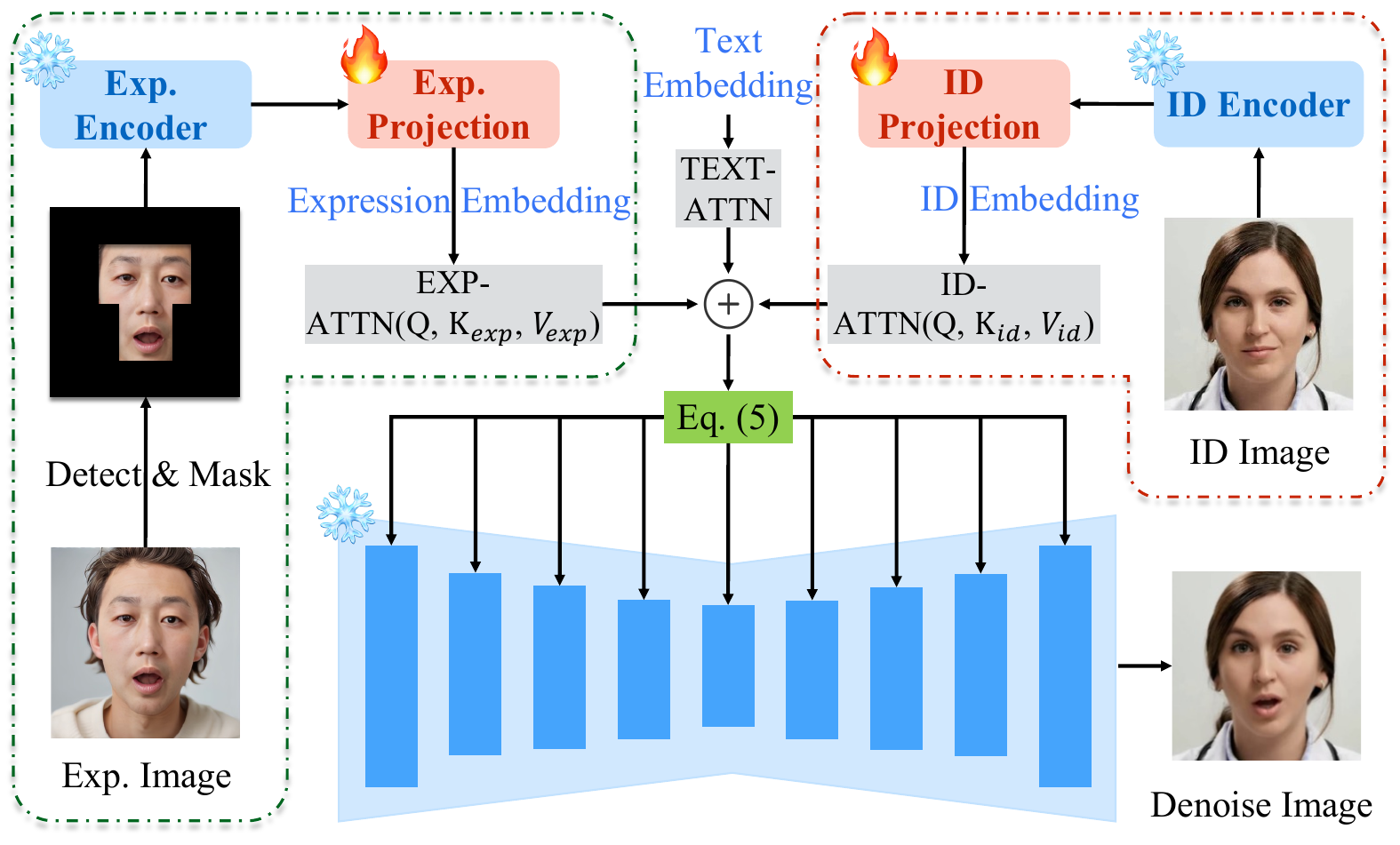}
    \caption{\textbf{The proposed E-Adapter}. The embeddings of identity and expression images are obtained through respective branches. Subsequently, the ID embedding, text embedding, and expression embeddings are integrated into the network, as depicted in Eq. (\ref{eq:eq6}).} 
    \label{fig:adapter}
  \end{figure}

\section{Related Work}
\label{sec:related}
\subsection{ID-Preserving Image Generation}
The task of identity-preserving generation \cite{ruiz2023dreambooth, gal2022inversion, hu2021lora} seeks to empower diffusion models to synthesize
images of given identities. Recent studies \cite{li2024photomaker,chen2023photoverse,huang2024consistentid,guo2024pulid} have explored tuning-free methods, which utilize face embeddings to conditioning generation without altering raw models. For instance, FaceChain \cite{liu2023facechain} introduces an independent adapter before the cross-attention layer, preventing interference between face and text conditions. IP-Adapter \cite{ye2023ipa} introduces a novel decoupled cross-attention mechanism for identity injection. InstantID \cite{wang2024instantid} proposes an IdentityNet to incorporate facial key points with image conditions. Although these methods can quickly generate customized portraits, they severely lack fine control over the expressions.

\subsection{Expression Customization}
Most methods for expression animation integrate control signals extracted from expression images into diffusion models. ControlNet \cite{zhang2023controlnet} is a pioneering effort that enables the incorporation of human poses, depth maps, and other image conditions. Following this framework, MediaPipe \cite{MediaPipeFace} specializes in extracting facial landmarks from reference expressions to preserve more expression details. Arc2Face \cite{deng2019arcface} extracts 3D normal maps \cite{filntisis2022emoca} to personalize expressions based on given portraits. Additionally, some studies pursue more flexible control. Collaborative Diffusion \cite{huang2023collaborative} allows localized editing of facial parts using semantic maps. \cite{paskaleva2024unified} combined valence \& arousal, and action units and GANmut \cite{d2021ganmut} to characterize emotion information. DiffSFSR \cite{liu2024DiffSFSR} utilizes an emotional dictionary to transfer facial expressions to the generated image. FineFace \cite{varanka2024fineface} employs action units to allow localized control by describing the intensity of facial muscle movements.

In practice, portrait animation \cite{zeng2023fadm,hu2024animateanyone,ma2024followyouremoji,drobyshev2024emoportraits,chen2024echomimic,wang2024instructavatar,wei2024aniportrait} can adjust portrait expressions without altering the image background. 
Notable examples, such as  FaceAdapter \cite{han2024faceadapter}, X-Portrait \cite{xie2024xportrait}, and LivePortrait \cite{guo2024liveportrait}, which can be used for post-processing of generated portrait images.
However, these methods cannot simultaneously customize real-world identities and expressions, adversely affecting identity fidelity and image quality.

\section{Methodology}
\label{sec:method}
Given a source portrait $\mathbf{S}$, a textual prompt $\mathbf{P}$, and a reference $\mathbf{R}$, our method aims to generate target images $\mathbf{T}$. To enable the model to accept RGB expression inputs, we aim to construct ternary data sets $\mathcal{T}=\{\mathbf{S}_i^{\neg{e}}, \mathbf{R}_{\neg{i}}^e, \mathbf{T}_i^e\}$ for decoupled training. Here, $i$ and $e$ denote the specific identity and the expression, and $\neg{i}$ and $\neg{e}$ can be any specific items different from $i$ and $e$. During training, the model takes $\mathbf{S}_i^{\neg{e}}, \mathbf{R}_{\neg{i}}^e$ as inputs and $\mathbf{T}_i^e$ as the learning objective, represented as $\mathbf{T}_i^e=\mathcal{F}(\mathbf{S}_i^{\neg{e}}, \mathbf{R}_{\neg{i}}^e)$, where $\mathcal{F}$ is the trained expression controller. This setup allows the model to maintain the identity $i$ of $\mathbf{S}$ while accurately replicating the expression $e$ from $\mathbf{R}$.
However, we are unable to collect qualified cross-identity expression pairs $\{\mathbf{R}_{\neg{i}}^e, \mathbf{T}_i^e\}$ from real-world scenarios. Besides, the integration of expression control hinders the original ID control.

To overcome these issues, our method is divided into two stages: decoupling training and fine-tuning.
Before decoupled training, we construct the required data delicately in two steps. Firstly, we design the expression controller and train a base version (Base E-Adapter) using the same-identity ternary data sets as detailed in Sec. \ref{sec:base E-Adapter}. The images generated by Base E-Adapter exhibit a very high expression consistency with given expression references, although they also have a similar appearance to the references (ID leakage).
Secondly, we develop the ID-irrelevant Data Iteration (IDI) to alter the identity of the images generated by Base E-Adapter, thus promoting the given reference expressions and the altered images to form data pairs with the same expression but different identity (as outlined in Sec. \ref{sec:iteration}).
Using the newly created cross-identity dataset, we perform decoupled training on the proposed expression controller to develop the Refined E-Adapter. Refined E-Adapter accurately controls expression details while preventing ID leakage when generating portrait images (see Sec. \ref{sec:refined E-Adapter}).
Finally, we introduce ID-enhanced Contrast Alignment (ICA) to further fine-tune the Refined E-Adapter, reducing the negative impact of expression manipulation on ID control (Sec. \ref{sec:contrast alignment}).
Fig. \ref{fig:pipeline} provides an overview of our pipeline.

\subsection{Preliminary}
\label{sec:preliminary}

\textbf{Latent Diffusion Model.} The latent diffusion models involve a diffusion process and a reverse process in the latent space. During the diffusion process, an image $x$ is first projected to a smaller latent representation by a VAE \cite{kingma2013vae} encoder. Then, random noise is gradually added to the representation $z_0$. The noisy representation $z_t$ is derived as:
\begin{equation}\label{eq:eq1}
 z_t=\sqrt{\bar\alpha_t}z_0+\sqrt{1-\bar\alpha_t}\epsilon, 
  \end{equation}
where $t$, $\bar\alpha_t$ denotes timestep and predefined function of $t$, respectively.
Conversely, during the denoising process, a U-Net \cite{ronneberger2015unet} is trained to predict the added noise $\epsilon$ from the noisy latent representation $z_t$. The training objective aims to minimize the loss function as follows:
\begin{equation}\label{eq:eq2}
\mathcal{L}_{sd}=E_{z_0,\epsilon\sim\mathcal{N}(\mu,\sigma^2)}\left[\left\|\epsilon-\epsilon_{\theta}(z_t,t,C)\right\|_2^2\right],
\end{equation}
where $C$ represents the condition. 

\textbf{Image Prompt Adapter. } IP-Adapter \cite{ye2023ipa} introduces a novel approach to incorporating an image prompt, working together with a text prompt to control image generation. A decoupled cross-attention is used to separate cross-attention layers for text features and image features, formulated as:
\begin{equation}\label{eq:eq3}
  \begin{aligned}
 Z={\rm Attn}(Q, K_t, V_t)+\lambda\cdot{\rm Attn}(Q, K_i, V_i),
  \end{aligned}
\end{equation}
where $\lambda$ is a weight factor, and $\{Q, K, V\}_{source}$ refers to the source of the feature tensor. 

\subsection{Decoupled Training}
Cross-identity expression pairs $\{\mathbf{R}_{\neg{i}}^e, \mathbf{T}_i^e\}$ are essential for training expression controllers. However, we cannot obtain adequate samples from the real world due to the subtle differences in individual expressions. Additionally, existing methods \cite{guo2024liveportrait, FaceFusion, xie2024xportrait,yang2024megactor} also struggle to synthesize qualified data pairs, particularly with extreme expressions. Therefore, we seek to develop a novel strategy for data construction. \label{dis:IDI} Reviewing previous works \cite{xie2024xportrait,yang2024megactor}, we notice that they avoid using same-identity expression pairs $\{\mathbf{R}_i^e, \mathbf{T}_i^e\}$ for controller training to prevent identity leakage during inference.
However, our insights suggest that \textbf{the controller trained with same-identity pairs exhibits superior expression transfer capabilities}, as shown in Fig. \ref{fig:iter}. 
Based upon this, we divide the synthesis of cross-identity expression pairs into two sequential steps: maintaining expression and modifying identity.
Initially, we utilize the same-identity data to train an expression controller that accurately transfers reference expressions into generated images without considering identity leakage. Subsequently, we utilize the portrait generation process to further alter the identity of the controller-generated image.
With the powerful abilities of the diffusion model, the modified image and the given expression template form a high-quality cross-identity data pair. After that, the data can be used to train the model to decouple expressions from identities.

\subsubsection{Base E-Adapter Training}
\label{sec:base E-Adapter}
In this subsection, we first propose the expression controller (E-Adapter) suitable for portrait generation and then train its basic version for subsequent data synthesis. 
Given the strong coupling between identity and expression, we employ a decoupled parallel attention to integrate them and design the E-Adapter, building upon IP-Adapter \cite{ye2023ipa}. As illustrated in Fig. \ref{fig:adapter}, apart from text input, E-Adapter includes two control branches: the identity and expression branch. This simple yet effective design achieves three objectives. First, the parallel control of expression and identity reduces operation errors compared to the serial way. Second, the model can efficiently retrieve features from identity or expression, and learn how to incorporate the expression into the given identity. Third, the training module fully decouples expression and identity control, enabling independent or simultaneous manipulation of both (more empirical analysis can be found in Sec. \ref{exp:controller} and Sec. \ref{sec:generalizability}).
Specifically, in the identity branch, we inherited IP-Adapter to encode identity embeddings and inject them into the U-Net along with text embeddings, formulated as:
\begin{equation}\label{eq:eq5}
  \begin{aligned}
 Z_{id} = &{\rm Attn}(Q_{noise}, K_{text}, V_{text}) \\
    &+ \lambda_{id} \cdot {\rm Attn}(Q_{noise}, K_{id}, V_{id}),
  \end{aligned}
\end{equation}
where $\lambda_{id}$ refers to the strength of the ID control. 

In the expression branch, we first preprocess the original expression reference to obtain a local image, which removes irrelevant elements such as the background, face contour, etc.
We then extract the expression embedding and inject it into the same cross-attention module with the identity embedding, denoted as:
\begin{equation}\label{eq:eq6}
 Z_{exp} \ = Z_{id} + \lambda_{exp} \cdot {\rm Attn}(Q_{noise}, K_{exp}, V_{exp}),
\end{equation}
where $\lambda_{exp}$ refers to the strength of the expression control. 

Based on the designed controller, we utilize the same-identity data $\{\mathbf{S}_i^{\neg{e}}, \mathbf{R}_{i}^e, \mathbf{T}_i^e\}$ to train the Base E-Adapter, adopting denoising loss $\mathcal{L}_{sd}$ throughout the process.

\subsubsection{ID-irrelevant Data Iteration}
\label{sec:iteration}
\begin{figure}[!tp]
  \centering
  \includegraphics[width=0.85\columnwidth]{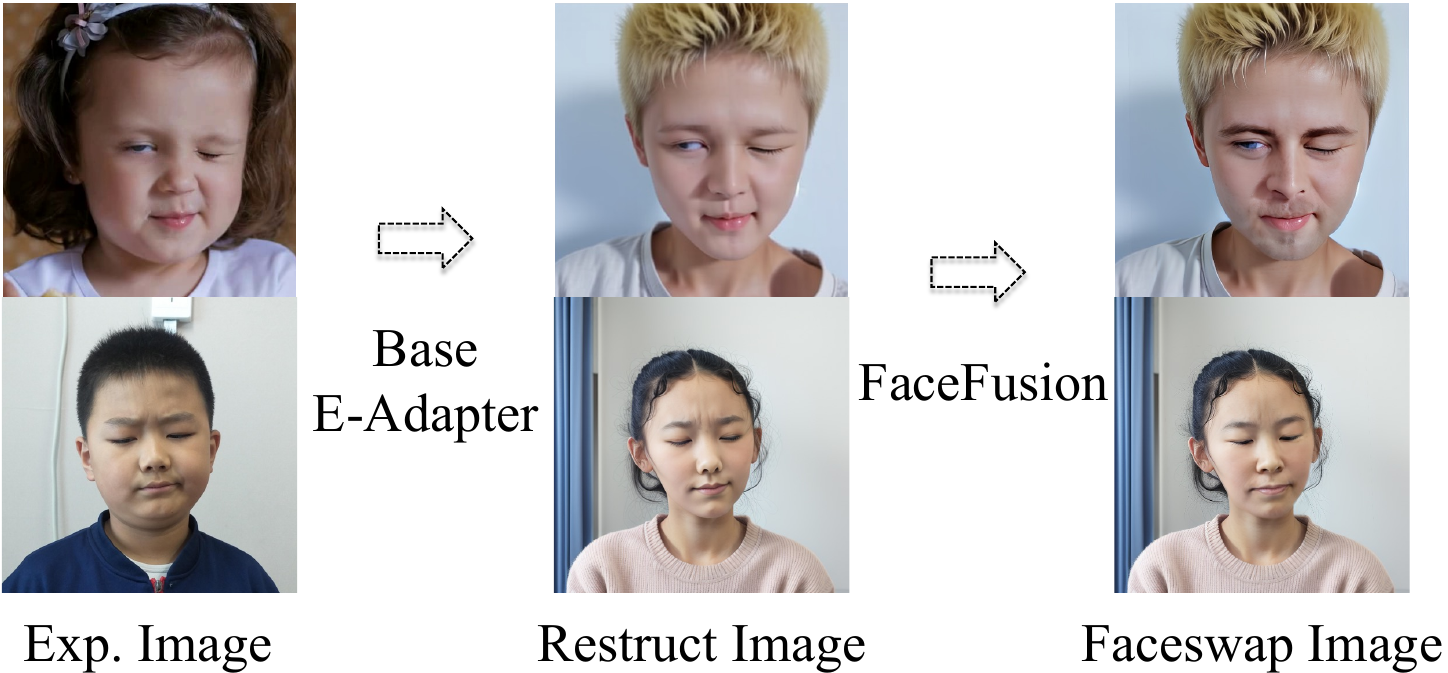}
  \caption{\textbf{Examples of IDI}, detailed in Sec. \ref{sec:iteration}}.
  \label{fig:iter}
\end{figure}

The image identity constructed using the Base E-Adapter is greatly influenced by the expression reference's appearance. The core idea of IDI is to minimize this disruption.
To achieve this, we innovatively reuse Base E-Adapter to modify the identity of the generated image while keeping the expression unchanged.
Specifically, we utilize expression images $\mathbf{R}_i^e$ inherited from training as the expression input but random identity $\neg{i}$ as the input of the identity branch, combining with the diffusion model to synthesize the restruct image. Subsequently, we perform FaceFusion \cite{FaceFusion} on the restruct image to create the faceswap image, further increasing the identity disparity between the generated image and the expression reference. Meanwhile, we employ facial blendshapes \cite{BlendshapeV2} and landmark \cite{lugaresi2019mediapipe} differences to exclude unqualified data (detailed in \textit{supplementary material} Sec. \ref{sec:sup_detail}). Through these processes, we acquire cross-identity expression pairs $\{\mathbf{R}_{\neg{i}}^e, \mathbf{T}_i^e\}$ that maintain highly consistent expressions while exhibiting completely distinct identity, and the example is illustrated in Fig. \ref{fig:iter}.

\textbf{Why can IDI keep the expression unchanged while changing the image's identity?}
On one hand, as our insights at the beginning of Sec. \ref{dis:IDI}, E-Adapter trained with the same identity data exhibits superior expression transfer capability, since the model tends to replicate the appearance and texture of the reference image. Additionally, the expression branch of the Base E-Adapter has been overfit to all expression images by training. Consequently, IDI can effortlessly transmit the expression into the restruct images.
On the other hand, the ID branch possesses certain identity control capabilities. By reusing the ID branch, we inject high-level semantic features of randomized portraits (\eg, gender, hairstyle, and face shape) into the restruct images. Notably, the identity of the generated image does not need to match the input ID condition precisely; it simply needs to differ from that of the expression condition. Moreover, the FaceFusion operation further alters the low-level facial textures, ensuring that the identity of the faceswap image is entirely distinct from the input.

\subsubsection{Refined E-Adapter Decoupled Training}
\label{sec:refined E-Adapter}
Utilizing the newly developed cross-identity data, we train a new expression controller termed the Refined E-Adapter. The controller's expression and ID branches accept input from different portraits respectively. Consequently, the identity and expression control branches are explicitly decoupled in their functions, each responsible solely for conveying identity and expression, respectively.

\subsection{ID-enhanced Contrast Alignment}
\label{sec:contrast alignment}
Both identity control and expression control manipulate the face of the generated image, causing potential interference between the two. Inevitably, incorporating expression control affects the original identity control in portrait generation tasks. Our findings indicate that \textbf{emphasizing expression control alters muscle details and facial attributes} (\eg, the presence of glasses, hair, and facial textures), due to the generative diversity inherent in the diffusion model.

To this end, we propose ICA to fine-tune the E-Adapter. 
Based on the same inputs, ICA incorporates the frozen Refined E-Adapter and the original identity controller (IP-Adapter \cite{ye2023ipa}) as supervision to calculate expression loss $\mathcal{L}_{exp}$ and identity loss $\mathcal{L}_{id}$. $\mathcal{L}_{id}$ encourages the IP-Adapter and the E-Adapter which integrates expression control to generate similar portraits, thereby minimizing the impact of expression control on identity fidelity. Additionally, $\mathcal{L}_{exp}$ ensures that fine-tuning does not compromise the existing expression control capability of the Refined E-Adapter.

Both losses are dependent on the denoised clean image $\hat{x}_0$. However, standard diffusion inference incurs significant gradient backpropagation time costs and GPU memory usage, and it will increase linearly with the number of timesteps.
To overcome this, we propose Adaptive Noise Inversion (ANI), an efficient one-step decoding strategy, denoted by the formula:
\begin{equation}\label{eq:eq7}
  \begin{aligned}
 \hat{x}_0 &= {\rm Decode}\left(z_0 + f(t) \cdot \left(\epsilon - \epsilon_{\theta}(z_t, t, C)\right)\right), \\
 f(t) &=
  \begin{cases}
 \sqrt{\frac{1-\bar{\alpha}_t}{\bar{\alpha}_t}} & \text{if } t \leq R_{tmax}, \\

 f(R_{tmax}) & \text{if } t > R_{tmax},
  \end{cases}
  \end{aligned}
  \end{equation}
where "${\rm Decode}$" is the VAE decode function, and $R_{tmax}$ is a constant. 
Through ANI, we can directly reconstruct images $\hat{x}_0$ at any given timestep $t$, even with large timesteps characterized by high noise.
We present \textbf{theoretical analysis} of this mechanism in \textit{supplementary material} Sec. \ref{sec:sup_ANI}.

Building on ANI, we employ the IP-Adapter to generate portrait image $\hat{x}_0^{id}$, and the currently fine-tuning E-Adapter to generate the image with expressions $\hat{x}_0^{cur}$. $\mathcal{L}_{id}$ minimizes the identity difference between two images, defined as:
\begin{equation}\label{eq:eq8}
  \begin{aligned}
 \mathcal{L}_{id} = 1 - \cos(\phi(\hat{x}_0^{id}), \phi(\hat{x}_0^{cur})),
  \end{aligned}
\end{equation} 
where $\cos$ is cosine similarity, and $\phi(\cdot)$ is the pre-trained face embedding extractor \cite{deng2019arcface}. Here, we use $\hat{x}_0^{id}$ rather than another clean portrait image as ground truth of $\mathcal{L}_{id}$, since we find that the noise discrepancy between the real image and one-step decoded image $\phi(\hat{x}_0^{cur})$ destabilizes fine-tuning (ablated in Sec. \ref{sec:ablation}).
Similarly, we utilize the frozen Refined E-Adapter to generate image $\hat{x}_0^{exp}$. The expression loss $\mathcal{L}_{exp}$ reduces the difference in facial landmarks between $\hat{x}_0^{exp}$ and $\hat{x}_0^{cur}$, is defined as follows:
$$
\mathcal{L}_{exp} = {\rm Wing}(\varphi(\hat{x}_0^{exp}), \varphi(\hat{x}_0^{cur})),
$$
where "${\rm Wing}$" denotes the wing loss \cite{feng2018wing}, and $\varphi(\cdot)$ is the pre-trained facial landmarks detector \cite{song2022gratis}.

The loss during fine-tuning is formulated as below:
\begin{equation}\label{eq:eq4}
  \begin{aligned}
 \mathcal{L}_{ft} = \mathcal{L}_{sd} + w_{id} \cdot \mathcal{L}_{id} + w_{exp} \cdot \mathcal{L}_{exp},
  \end{aligned}
\end{equation} 
where $w_{id}$ and $w_{exp}$ are the balanced hyper-parameters.

\begin{figure*}[!htbp]
  \centering
  \includegraphics[width=0.95\textwidth]{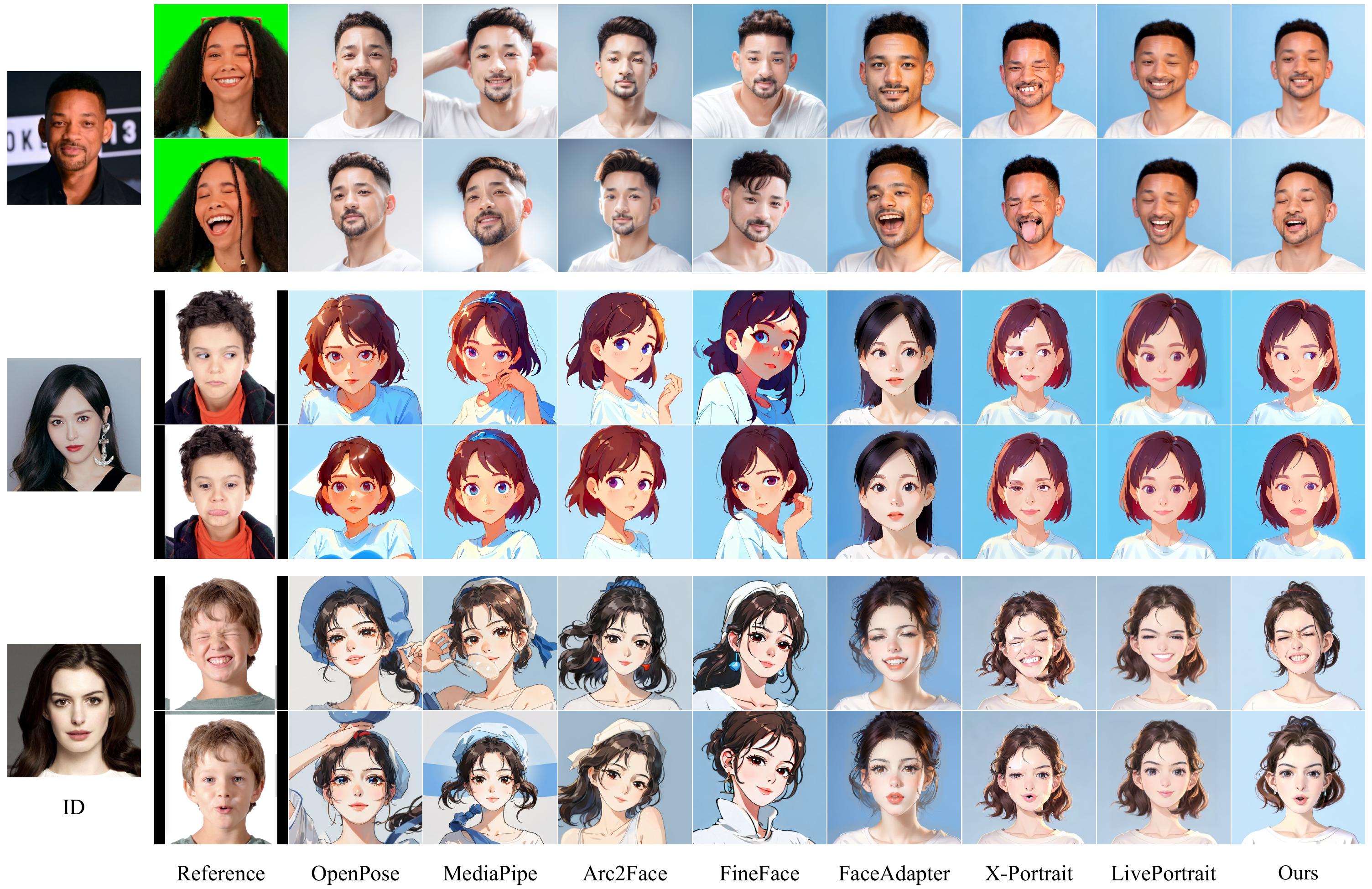}
  \caption{\textbf{Qualitative Comparisons}. Compared to other methods, our method demonstrates the most accurate and robust transfer of subtle facial expressions (\eg, pouting, single-eye blinks, and pupil movements) while preserving the identity of the source portrait, even when applied in artistic styles (such as anime and ink painting).
 } 
  \label{fig:compare}
  \vspace{-0.1cm}
\end{figure*}

\begin{table*}[]
  \centering
  \resizebox{0.85\textwidth}{!}{%
  \begin{tabular}{@{}lcccccccccccc@{}}
  \toprule
               & \multicolumn{4}{c}{\textbf{Realistic}}          & \multicolumn{4}{c}{\textbf{Anime}}        & \multicolumn{4}{c}{\textbf{Painting}}           \\ 
               \cmidrule(l){2-5} \cmidrule(l){6-9} \cmidrule(l){10-13}
  \multirow{-2}{*}{\textbf{Method}} &
    \textbf{ID $\uparrow$} &
    \textbf{IQ $\uparrow$ } &
    \textbf{Exp. $\downarrow$} &
    \textbf{LMS $\downarrow$} &
    \textbf{ID $\uparrow$} &
    \textbf{IQ $\uparrow$} &
    \textbf{Exp. $\downarrow$} &
    \textbf{LMS $\downarrow$} &
    \textbf{ID $\uparrow$} &
    \textbf{IQ $\uparrow$} &
    \textbf{Exp. $\downarrow$} &
    \textbf{LMS $\downarrow$} \\ \midrule
  \multicolumn{13}{l}{\cellcolor[HTML]{E7E9E8}Portrait Animation  }                                                                                               \\
 FaceAdapter \cite{han2024faceadapter}  & 0.309       & {\ul 4.991} & 0.059       & 0.281 & 0.095       & {\ul 4.853} & 0.109 & 0.467 & 0.146       & 4.747       & {\ul 0.072} & 0.348 \\
 X-Portrait \cite{xie2024xportrait}&
 0.505 &
 4.902 &
 {\ul 0.058} &
 {\ul 0.266} &
 0.188 &
 4.142 &
    \textbf{0.066} &
    \textbf{0.335} &
 0.332 &
 3.668 &
    \textbf{0.062} &
 {\ul 0.291} \\
 LivePortrait \cite{guo2024liveportrait}& 0.609       & 4.772       & {\ul 0.058} & 0.319 & {\ul 0.262} & 4.843       & 0.116 & 0.551 & {\ul 0.419} & {\ul 4.749} & 0.087       & 0.426 \\ \midrule
  \multicolumn{13}{l}{\cellcolor[HTML]{E7E9E8}Diffusion Conditioning}                                                                                          \\
 OpenPose  \cite{zhang2023controlnet}   & {\ul 0.628} & 4.981       & 0.097       & 0.447 & 0.162       & 4.741       & 0.142 & 0.698 & 0.326       & 4.484       & 0.121       & 0.507 \\
 MediaPipe \cite{lugaresi2019mediapipe}   & 0.623       & 4.985       & 0.104       & 0.472 & 0.087       & 4.713       & 0.149 & 0.671 & 0.287       & 4.476       & 0.130       & 0.524 \\
 Arc2Face \cite{deng2019arcface}    & 0.618       & 4.946       & 0.116       & 0.537 & 0.097       & 4.630       & 0.163 & 0.713 & 0.146       & 4.349       & 0.141       & 0.609 \\
 FineFace \cite{varanka2024fineface}   & 0.532       & 4.870       & 0.120       & 0.576 & 0.090       & 4.423       & 0.156 & 0.700 & 0.297       & 3.813       & 0.142       & 0.624 \\ 
  \cmidrule(l){2-13} 
 EmojiDiff (Ours) &
    \textbf{0.666} &
    \textbf{4.995} &
    \textbf{0.054} &
    \textbf{0.215} &
    \textbf{0.304} &
    \textbf{4.910} &
 {\ul 0.095} &
 {\ul 0.359} &
    \textbf{0.469} &
    \textbf{4.756} &
 0.078 &
    \textbf{0.256} \\ \bottomrule
  \end{tabular}%
 }
  \caption{
    Quantitative comparisons of our method with SOTA counterparts in various base models. \textbf{Bold} and {\ul{underlined}} values denote the best and second-best results respectively.
    }
  \label{tab:main}
  \vspace{-0.1cm}
  \end{table*}

\section{Experiments}
\label{sec:exp}
\subsection{Experiment Setting}
\textbf{Dataset.} We first collect 10k indoor and outdoor images of people of different countries and genders, each with 3 to 12 expressions. Then, we use LIQE \cite{zhang2023liqe} to filter out low-quality images and crop out 512$\times$512 face square images. Subsequently, this dataset is employed to train the Base E-Adapter based on SD1.5 \cite{rombach2022sd}. Afterward, we construct about 120k cross-identity triplets through IDI. Through expression consistency filtering, we retain about 100k triplets named \textbf{CIEP100k}, for training and fine-tuning the Refined E-Adapter. For evaluation, we gather images of 50 celebrities from diverse fields, with 30 different expressions.
        
\textbf{Implementation Details.} We train the Base E-Adapter on SD1.5 to construct data and the Refined E-Adapter on both SD1.5 \cite{rombach2022sd} and SDXL \cite{podell2023sdxl}.
In the Base E-Adapter training stage, our model is trained from scratch for 0.5 days. 
In SD1.5, we train the Refined E-Adapter for 1 day and fine-tune it for 4 hours. In SDXL, we train the Refined E-Adapter for 2 days and fine-tune it for 0.5 days. More details are provided in \textit{supplementary material} Sec. \ref{sec:sup_detail}.


\textbf{Baselines.} We compare our model with prior methods, including OpenPose \cite{zhang2023controlnet}, MediaPipe \cite{lugaresi2019mediapipe}, Arc2Face \cite{paraperas2024arc2face}, and FineFace \cite{varanka2024fineface}, using the same configurations (base model, text prompt, inference steps, etc.). 
Although portrait animation methods are fundamentally different from our approach, we assess these methods, including FaceAdapter \cite{han2024faceadapter}, X-Portrait \cite{xie2024xportrait}, and LivePortrait \cite{guo2024liveportrait}, to thoroughly evaluate our method.
For the diffusion-based methods, we integrate IP-Adapter \cite{ye2023ipa} plugin for identity preservation. 
For animation-based methods, an expressionless image of the portrait is generated, and then animation is applied.
We select LivePortrait \cite{guo2024liveportrait} to construct data triples $\mathcal{T}$ and train the E-Adapter, which serves as our strong baseline.

\textbf{Benchmarks.}        
For a fair comparison, all experiments are conducted using the SD1.5 framework across three styles: realistic, anime, and ink painting, while the style is directly dependent on the base model and style LoRA.
Given the identity image, expression reference, and the generated image, we evaluate all methods using ID fidelity (ID), image quality (IQ), expression similarity (Exp.), and facial landmark movements similarity (LMS). 
Different from ID encoder in E-Adapter, the Antelopev2 \cite{Antelopev2} model is utilized to extract identity embeddings from both the portrait and generated image, with the cosine similarity between them used to evaluate identity fidelity. 
For the assessment of image quality, the pre-trained network LIQE \cite{zhang2023liqe} is employed. 
Expression similarity is measured by computing the L1 difference between facial blendshapes extracted from the expression and the generated image using MediaPipe \cite{lugaresi2019mediapipe}.
To evaluate the keypoint similarity, crucial facial landmarks (such as eyes, pupils, and mouth) are detected \cite{lugaresi2019mediapipe} to calculate movement amplitude differences. 
More metric details are provided in \textit{supplementary} Sec. \ref{sec:sup_metric}.

\subsection{Comparison }
\label{sec:comparison}
\textbf{Quantitative results.}
As summarized in Tab. \ref{tab:main}, our method consistently outperforms all competitors by a good margin.
Compared to the same type of diffusion conditioning methods, our approach shows significant improvements in ID similarity, image quality, and expression control ability. 
Even evaluating against portrait animation methods such as FaceAdapter \cite{han2024faceadapter} and LivePortrait \cite{guo2024liveportrait}, the proposed method offers superior ID fidelity and more accurate expression control. Notably, X-Portrait, which directly uses RGB expression images, also demonstrates strong expression control. However, {X-Portrait shows a noticeable drop in image quality when applied to non-realistic styles, with IQ scores of only 4.142 and 3.668 in the anime and painting styles, respectively.} On the contrary, our method produces exquisite stylized images due to the seamless integration with the diffusion model. We provide \textbf{comparisons with more methods} in the \textit{supplementary material} Sec. \ref{sec:sup_compare}.

\textbf{Qualitative results.}
Fig. \ref{fig:compare} illustrates the results of various methods across three styles. These examples demonstrate that our model effectively conveys reference expressions, including lip movements and eye gaze while preserving the portrait identity. The fourth and sixth examples in the figure highlight the ability of our method to capture subtle movements, including lower lip pursing and pouting. 
Similarly, as illustrated in Fig. \ref{fig:ICA}, the complete method retains more identity details(\eg, glasses and hair color), compared to the method without ICA.

\subsection{Ablation Study}
\label{sec:ablation}
We evaluate the proposed IDI and ICA on SD1.5 and SDXL. As presented in Tab. \ref{tab:module}, in comparison to our baseline, IDI substantially improves expression control, reducing the Exp. score of the two base models by 23.0\% and 22.6\%, respectively. Additionally, ICA further enhances ID fidelity without compromising expression control, increasing the corresponding ID score by 3\% and 6\%, respectively.

\begin{table}[!t]
  \centering
  \resizebox{0.88\columnwidth}{!}{%
  \begin{tabular}{@{}cccccccc@{}}
  \toprule
  \multirow{2}{*}{\textbf{IDI}} &
    \multirow{2}{*}{\textbf{ICA}} &
    \multicolumn{3}{c}{\textbf{SD1.5 \cite{rombach2022sd}}} &
    \multicolumn{3}{c}{\textbf{SDXL \cite{podell2023sdxl}}} \\ \cmidrule(l){3-5} \cmidrule(l){6-8} 
    &   & \textbf{ID $\uparrow$} & \textbf{Exp. $\downarrow$} & \textbf{LMS $\downarrow$} & \textbf{ID $\uparrow $} & \textbf{Exp. $\downarrow$} & \textbf{LMS $\downarrow$} \\ \midrule
    &   & 0.265       & 0.113         & 0.447        & 0.432       & 0.093         & 0.454        \\
  $\surd$ &   & 0.295       & 0.087         & 0.355        & 0.465       & 0.072         & 0.338        \\
  $\surd$ & $\surd$ & 0.304       & 0.095         & 0.359        & 0.509       & 0.071         & 0.341        \\ \bottomrule
  \end{tabular}%
 }
  \caption{Quantitative ablation about IDI and ICA.}

  \label{tab:module}
  \end{table}

\textbf{ID-irrelevant Data Iteration.} As shown in Tab. \ref{tab:data iteration}, we assess the performance of E-Adapter at different data iteration stages. Initially, the Base E-Adapter achieves significant reductions in Exp. and LMS scores compared to the baseline, showing the powerful expression transfer ability of Base E-Adapter. After using the constructed restruct images, the newly trained Refined E-Adapter substantially reduces ID leakage and improves the ID score by 0.136. Finally, through further face-swapping on the data, the Refined E-Adapter significantly outperforms the baseline.

\textbf{ID-enhanced Contrast Alignment.} As shown in Tab. \ref{tab:abl_ICA},
when utilizing clean real images as supervision, the ANI effectively improves ID fidelity and expression controllability. Additionally, using model predictions $\hat{x}_0^{id}$ rather than real images ensures consistency of noise levels between data pairs, further facilitating model fine-tuning.

\textbf{Expression Controller Architecture.}
\label{exp:controller}
We experiment with various expression controller structures (The detailed structure can be found in the \textit{supplementary material} Sec. \ref{sec:sup_controller}.). As indicated in Tab. \ref{tab:arch}, our approach combines identity and expression embeddings within a unified cross-attention block, effectively fostering their interaction and delivering optimal performance. 

\subsection{Generalizability of Proposed Approach}
\label{sec:generalizability}
\begin{figure}[!t]
  \centering
  \includegraphics[width=1\columnwidth]{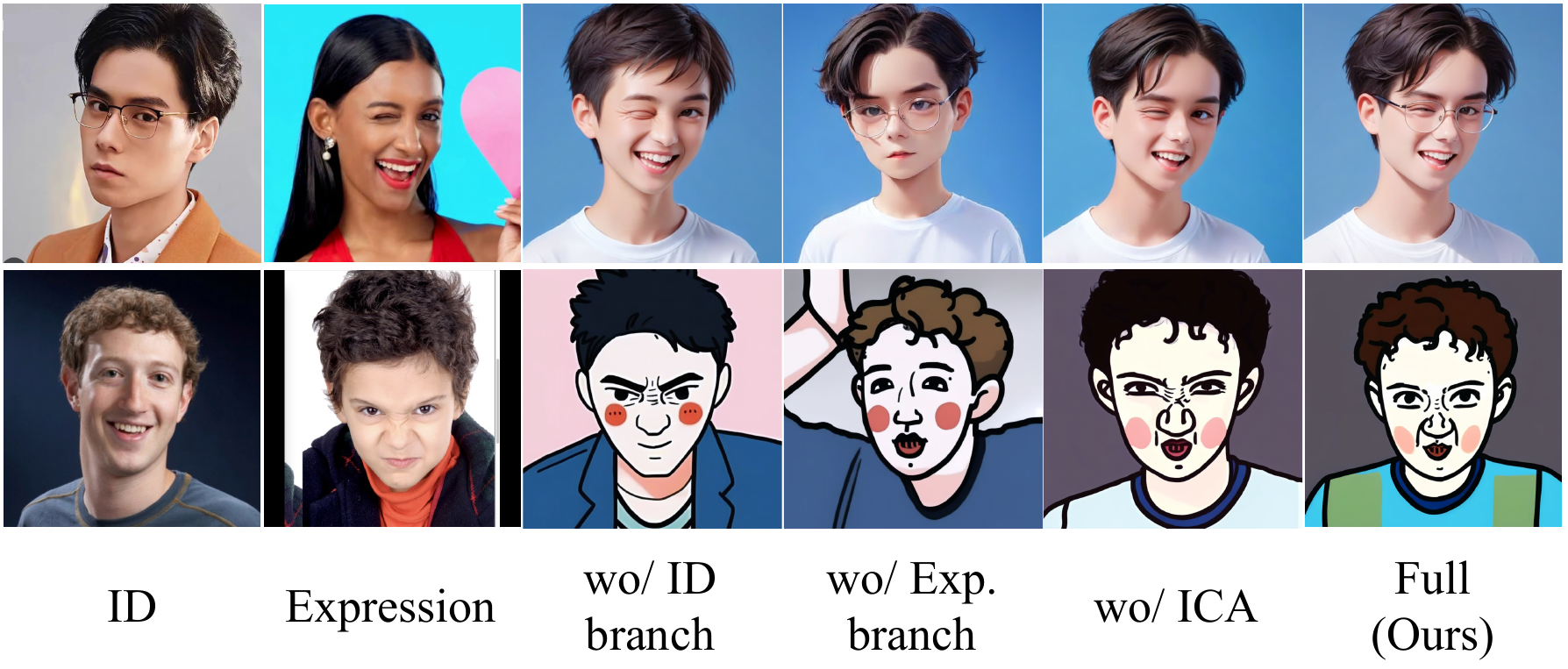}
  \caption{{Images generated without using the ID branch, expression branch, and ICA}.} 
  \label{fig:ICA}
\end{figure}

\textbf{Decoupled identity and expression.} As shown in Fig. \ref{fig:ICA}, the third and fourth columns demonstrate that the identity and expression branch can take effect independently, enhancing the flexibility of facial control.

\textbf{E-Adapter as the expression extractor}. The expression embedding from our method can be generalized to expression-related tasks. Specifically, we evaluate the expression module of the E-Adapter on the expression recognition task following the classic method DLN \cite{Zhang2021Learning}. Our method exhibits powerful expression recognition capabilities (88.7 \textit{vs} DLN’s 86.4 in the RAF-DB benchmark \cite{RAF}), demonstrating its efficacy as a robust expression extractor.

\textbf{Generalization to other architecture}. 
In the main experiment, considering generalizability of method and fair comparison with the SD1.5-based animation methods (\eg, XPortrait \cite{xie2024xportrait}, FaceAdapter \cite{han2024faceadapter}), we simply injected condition information via the attention mechanism, which is similar to IP-Adapter \cite{ye2023ipa}.
As shown in Tab. \ref{tab:compare_portrait}, by directly utilizing the CIEP100k dataset and ICA, our method is seamlessly adopted to InstantID \cite{wang2024instantid} and achieves substantially more detailed expression control.

  \begin{table}[!t]
    \centering
    \resizebox{0.88\columnwidth}{!}{%
    \begin{tabular}{lccc}
    \toprule
    \textbf{Stage / Data Source} & \textbf{ID $\uparrow$} & \textbf{Exp. $\downarrow$} & \textbf{LMS $\downarrow$} \\ \midrule
   Baseline                   & 0.265 & 0.113 & 0.447        \\
   Base E-Aadapter                & 0.127 & 0.065 & 0.283        \\
   using Restruct Image                  & 0.263 & 0.080 & 0.362        \\
   Ours (using Faceswap Image)                 & 0.295 & 0.087 & 0.355         \\ \bottomrule
    \end{tabular}%
   }
    \caption{Ablation study of IDI.}
  
    \label{tab:data iteration}
  \end{table}

  \begin{table}[]
    \centering
    \resizebox{0.68\columnwidth}{!}{%
    \begin{tabular}{@{}llll@{}}
        \toprule
        {Supervision} & \textbf{ID $\uparrow$} & \textbf{Exp. $\downarrow$} & \textbf{LMS $\downarrow$}   \\ \midrule
        Clean Image           & 0.477 & 0.088 & 0.376 \\
        Clean Image w/ ANI          & 0.485 & 0.078 & 0.361 \\
        Ours           & \textbf{0.509} & \textbf{0.071} & \textbf{0.341} \\ \bottomrule
        \end{tabular}%
      }
      
      \caption{Ablation study of ICA supervision.}
      \label{tab:abl_ICA}
  \end{table}

\begin{table}[]
  \centering
  \resizebox{0.78\columnwidth}{!}{%
  \begin{tabular}{lccc}
  \toprule
  \textbf{Architecture} & \textbf{ID $\uparrow$} & \textbf{Exp. $\downarrow$} & \textbf{LMS $\downarrow$} \\ \midrule
 (a) SelfAttn                   & 0.266       & 0.109         & 0.497        \\
 (b)  ControlNet                & 0.294       & 0.121         & 0.552        \\
 (c) Sequential CrossAttn                  & 0.276       & 0.104         & 0.435        \\
 Ours                  & \textbf{0.295}       & \textbf{0.087}         & \textbf{0.355}        \\ \bottomrule
  \end{tabular}%
 }
  \caption{Ablation study of expression controller architecture.}

  \label{tab:arch}
\end{table}

\begin{table}[]
  \centering
  \resizebox{0.66\columnwidth}{!}{%
  \begin{tabular}{@{}llll@{}}
    \toprule
    {Method}                    & \textbf{ID $\uparrow$} & \textbf{Exp. $\downarrow$} & \textbf{LMS $\downarrow$} \\ \midrule
    PuLID \cite{guo2024pulid}                             & 0.536       & 0.117         & 0.502        \\
    PhotoMaker \cite{li2024photomaker}                             & 0.467       & 0.136         & 0.618        \\
    InstantID \cite{wang2024instantid}                         & 0.543       & 0.105         & 0.476        \\
    \cdashline{1-4}
    Ours                          & 0.509       & 0.071         & 0.341        \\
    Ours w/ InstantID & 0.540       & 0.077        & 0.356      \\ 
    \bottomrule
    \end{tabular}%
    }
    \caption{Comparisons with portrait generation methods.}
    \label{tab:compare_portrait}
\end{table}

\section{Conclusion}
\label{sec:conclusion}
  We introduce EmojiDiff, an innovative solution that utilizes both portrait images and RGB expression images to generate expressive portraits.
  To achieve this, we design a two-stage scheme including decoupling training and fine-tuning.
  In the training stage, we propose ID-irrelevant Data Iteration to synthesize identity-expression decoupling data pairs CIEP100k, enabling the model to accept the RGB expression template without interference from expression-irrelevant features.
  In the fine-tuning stage, we propose ID-enhanced Contrast Alignment to further harmonize identity and expression control. This approach ensures the training process through Adaptive Noise Inversion and mitigates the impact of expression modulation on identity fidelity.
  Experimental results demonstrate that our approach achieves accurate and nuanced facial expression control while maintaining precise identity resemblance.
\clearpage
{
    \small
    \bibliographystyle{ieeenat_fullname}
    \bibliography{main}
}

\clearpage
\renewcommand\thesection{\Alph{section}}
\setcounter{page}{1}
\setcounter{section}{0}
\setcounter{subsection}{0}
\setcounter{figure}{0}
\setcounter{table}{0}
\setcounter{equation}{0}

\maketitlesupplementary
In this supplementary material, we provide more experimental details in Sec. \ref{sec:exp}, 
detailed derivation of Adaptive Noise Inversion (ANI) in Sec. \ref{sec:sup_ANI}, additional experimental results in Sec. \ref{sec:results}.

\section{More Experimental Details}\label{sec:exp}
\subsection{Detailed Implementation}
\label{sec:sup_detail}
We gather images of over 10,000 people displaying various expressions from video clips and in-house face databases, removing low-quality images using LIQE \cite{zhang2023liqe}. Then, we employ the proposed IDI to construct cross-identity, same-expression datasets. 
As mentioned in Sec. 3.2.2 of the main body, 
facial blendshapes \cite{BlendshapeV2} and landmark \cite{lugaresi2019mediapipe} differences are utilized to filter out expression-changed data during data construction. Specifically, we calculate the Exp. and LMS metrics of the original expression image and the synthesized new image, excluding the data with Exp. $ \leq $0.05 and LMS $ \leq $0.18.
Based on synthesized data, we train the Refined E-Adapter for SD1.5 at a resolution of 512$\times$512. On SDXL, we upscale the data to a resolution of 1024$\times$1024 and train it with random scaling following IP-Adapter \cite{ye2023ipa}. The learning rates of the Adam optimizer during training and fine-tuning are set to 2$\times$ $10^{-5}$ and 5$\times$$10^{-6}$, respectively, with $\beta_1$ = 0.5, and $\beta_2$ = 0.999. In all experiments, $\lambda_{id}$, $\lambda_{exp}$, $R_{tmax}$, $w_{id}$, and $w_{exp}$ are consistently set to 1, 1, 600, 0.08, and 10, respectively.

In the E-Adapter, we employ CLIP \cite{clip} as the encoder for the expression branch to extract expression signals. The image-prompt and FaceID versions of IP-Adapter \cite{ye2023ipa} are utilized to initialize the projection layers of the expression and identity branches, emphasizing structure and identity features, respectively. During training and fine-tuning, the parameters of the expression branch and the projection layer of the identity branch are updated, while other parameters remain fixed. 
\subsection{Ablation Experiments Details}
\label{sec:sup_controller}
\begin{figure}[!t]
  \centering
  \includegraphics[width=0.9\columnwidth]{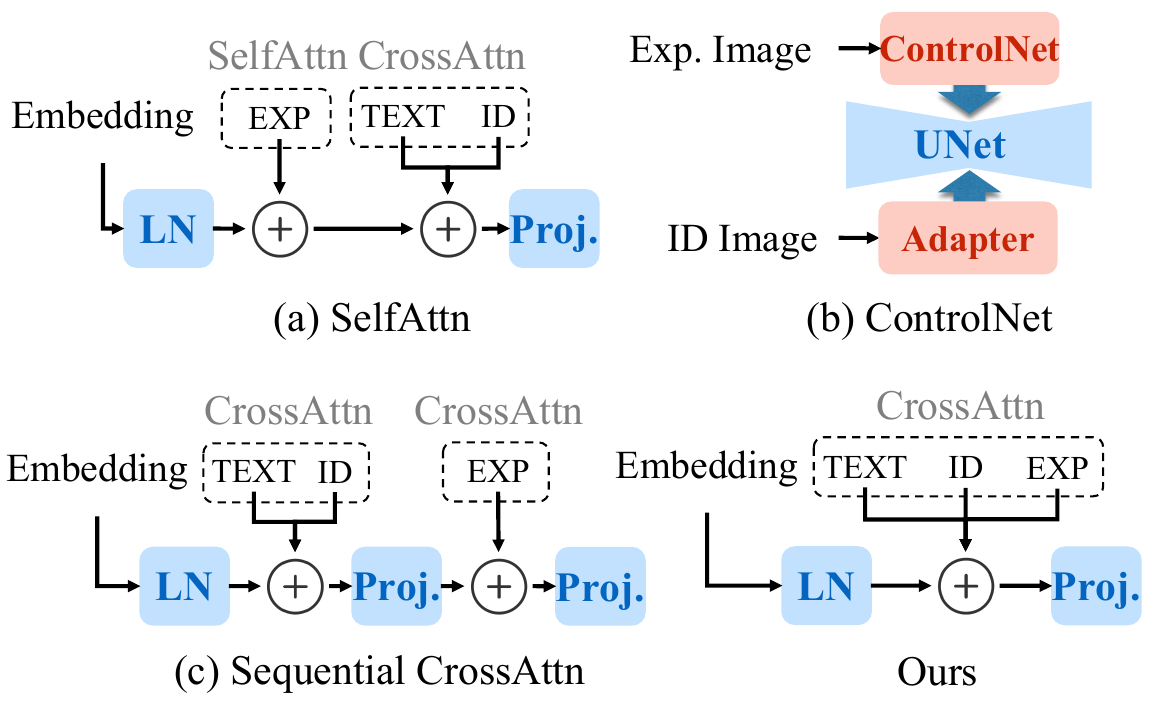}
  \vspace{-0.1cm}
  \caption{{Different expression controller structure}.} 
  \label{fig:arch}
\end{figure}

In Sec. 4.3 of the main body, we experiment with various expression controller structures. As is shown in the Fig. \ref{fig:arch}, expression embeddings are injected into the diffusion model using one of the following strategies: (a) self-attention, (b) ControlNet, (c) serial cross-attention with ID embeddings, and (d) parallel cross-attention with ID embeddings.

\subsection{Metric Details}
\label{sec:sup_metric}
\textbf{Exp.} We employ MediaPipe \cite{lugaresi2019mediapipe} to extract 52 facial blendshapes \cite{BlendshapeV2} of the expression reference and the generated image, represented as $b^{ref}$, $b^{gen}$, respectively. Each blendshape $b^{gen}_i$ in the [0, 1] range indicates the probability of the relevant facial action occurring (\eg, eyeBlinkLeft, mouthClose). The expression difference between reference and generated image can be formulated as:
\begin{equation}
  \begin{aligned}
 \mathbb{E}_{exp} = \sum_{i=0}^{51}\left(|b^{ref}_i - b^{gen}_i|\right)
  \end{aligned}
\end{equation}

\textbf{LMS} To further assess differences in the expression of critical facial regions (such as eyes, pupils, and mouth), we propose evaluating the landmark movement similarity (LMS) between the reference and generated image. First, we utilize MediaPipe \cite{lugaresi2019mediapipe} to detect 478 facial landmarks of the expression reference and generated image, denoted as $l^{ref}$ and $l^{gen}$, respectively. Next, we select representative landmarks to calculate the movement amplitude of key facial actions, including blinking, eye movement, and mouth opening, represented as:
\begin{equation}
  \begin{aligned}
 r_{leye} &=  \frac{\|l_{145}-l_{159}\|_{2}}{\max(10^{-5}, \|l_{133}-l_{33}\|_{2})},\\
 r_{lpupil} &=  \frac{\|l_{133}-l_{468}\|_{2}}{\max(10^{-5}, \|l_{133}-l_{33}\|_{2})},\\
 r_{reye} &=  \frac{\|l_{374}-l_{386}\|_{2}}{\max(10^{-5}, \|l_{263}-l_{362}\|_{2})},\\
 r_{rpupil} &=  \frac{\|l_{263}-l_{473}\|_{2}}{\max(10^{-5}, \|l_{263}-l_{362}\|_{2})},\\
 r_{mouth} &=  \frac{\|l_{17}-l_{0}\|_{2}}{\max(10^{-5}, \|l_{291}-l_{61}\|_{2})},\\
  \end{aligned}
\end{equation}
The LMS metric quantifies the difference in movement amplitude between two images, expressed as:

\begin{equation}\label{eq:eq5}
  \begin{aligned}
 \mathbb{E}_{LMS} = \sum_{k \in S}\left(|r_{k}^{ref} - r_{k}^{gen}|\right),
  \end{aligned}
\end{equation}
where $S=\{{\rm leye, reye, lpupil, rpupil, mouth}\}$.

\begin{figure}[!t]
  \centering
  \includegraphics[width=0.95\columnwidth]{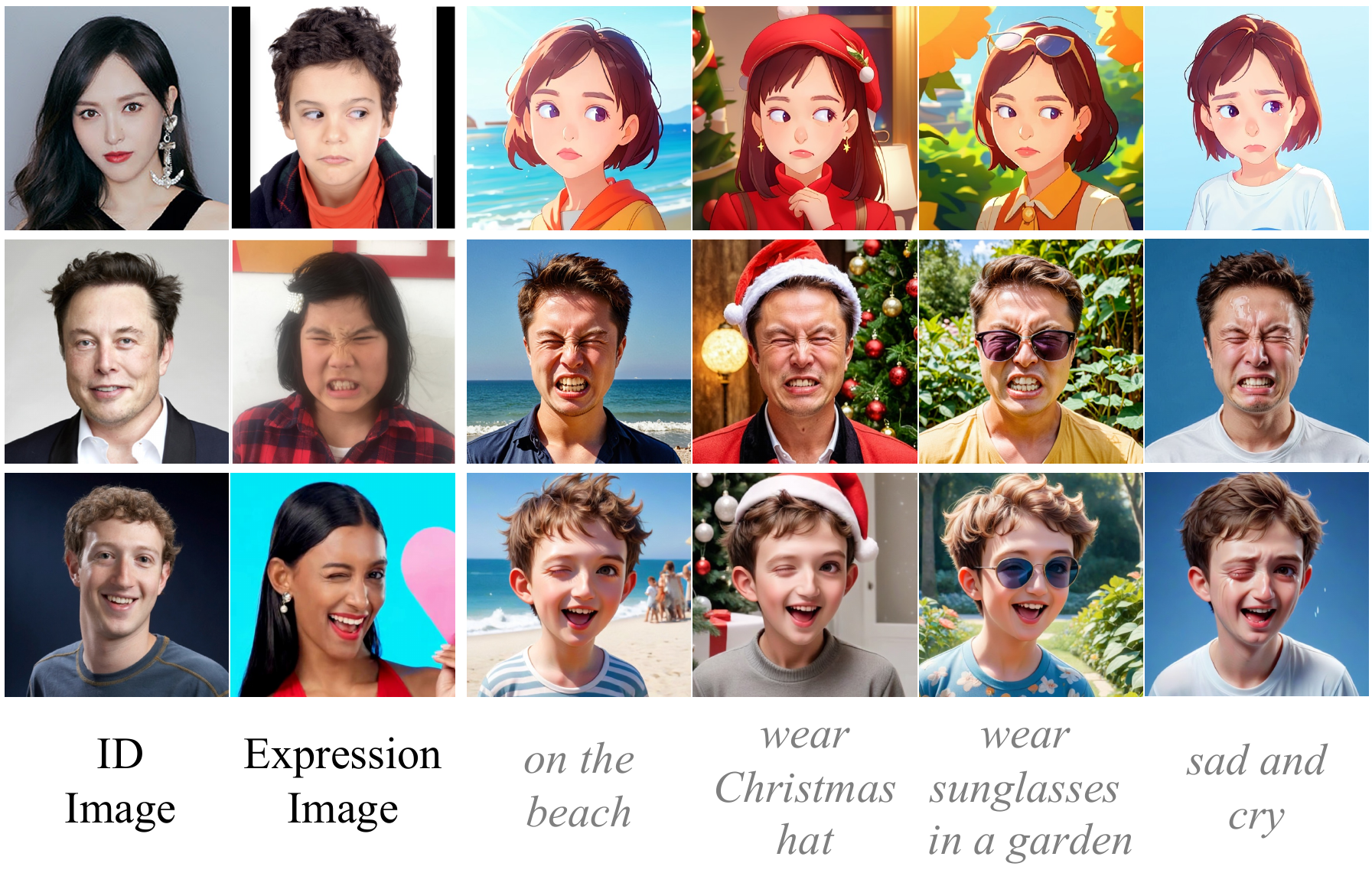}
  \caption{{Images generated using different prompts}.} 
  \label{fig:prompt}
\end{figure}
\begin{table}[]
  \centering
  \resizebox{0.65\columnwidth}{!}{%
  \begin{tabular}{@{}llll@{}}
    \toprule
    Method & \textbf{ID $\uparrow$} & \textbf{Exp. $\downarrow$} & \textbf{LMS $\downarrow$}   \\ \midrule
    DaGAN \cite{hong2022depth}         & 0.202 & 0.135 & 0.569 \\
    Follow-U-Emoji \cite{ma2024followyouremoji} & 0.243 & 0.125 & 0.428 \\
    AvatarArtist \cite{liu2025avatarartist}  & 0.239 & 0.110 & 0.454 \\
    SkyReels-A1 \cite{qiu2025skyreels}   & 0.277 & 0.104 & 0.387 \\
    Ours      & \textbf{0.304} & \textbf{0.095} & \textbf{0.359} \\ \bottomrule
    \end{tabular}%
  }
  \caption{Comparisons with post-processing animation methods.}
  \label{tab:sup_compare}
\end{table}

\section{Derivation of Adaptive Noise Inversion}
\label{sec:sup_ANI}
During the diffusion process, the noisy representation $z_t$ is derived from the original latent representation $z_0$ and added noise $\epsilon$, represented as:
\begin{equation}\label{eq:4}
  \begin{aligned}
 z_t={\sqrt{\bar{\alpha}_{t}}}z_0+\sqrt{1-\bar{\alpha}_t}*\epsilon,
  \end{aligned}
\end{equation}
where $t$, $\alpha_t$ represent timestep and a predefined function of $t$, respectively. 
When the noise perturbation is small, the noise  $\epsilon_{\theta}(z_t,t,C)$ predicted by U-Net approximately equals to the added noise $\epsilon$ \cite{peng2024portraitbooth,chen2023photoverse}.  During the denoising process, we can approximately reconstruct the original latent $z_0$ by performing single-step sampling, denoted as:
\begin{equation}\label{eq:5}
  \begin{aligned}
 \hat{z}_0 &= \frac{z_t-\sqrt{1-\bar{\alpha}_t}\ast \epsilon_{\theta}(z_t,t,C)}{\sqrt{\bar{\alpha}_{t}}},\\
\end{aligned}
\end{equation}
By combining the Eq. (\ref{eq:4}) and Eq. (\ref{eq:5}), the reconstructed latent $\hat{z}_0$ can also be expressed as:
\begin{equation}\label{eq:6}
  \begin{aligned}
 \hat{z}_0=z_0+\sqrt{\frac{1-\bar{\alpha}_t}{\bar{\alpha}_t}}\ast\left(\epsilon-\epsilon_{\theta}(z_t,t,C)\right)
  \end{aligned}
\end{equation}
As depicted in Eq. (\ref{eq:6}), the reconstructed latent $z_0$ can be directly derived from the origin latent $z_0$ and the difference between the added noise $\epsilon$ and predicted noise $\epsilon_{\theta}$, and the signal-to-noise ratio is decided by $\sqrt{\frac{1-\bar{\alpha}_t}{\bar{\alpha}_t}}$. 
The reconstructed sample $\hat{x}_0$ can be obtained by decoding $\hat{z}_0$, formulated as:
\begin{equation}\label{eq:7}
    \begin{aligned}
 \hat{x}_0&={\rm Decode}\left(z_0+\sqrt{\frac{1-\bar{\alpha}_t}{\bar{\alpha}_t}}\ast(\epsilon-\epsilon_{\theta}(z_t,t,C))\right),
    \end{aligned}
  \end{equation}
where "Decode" refers to the vae decode function. Notably, in the higher noisy stage (large timestep), the reconstructed sample $\hat{x}_0$ may become noisy, resulting in inaccurate identity and expression loss calculations. To overcome this, we propose the Adaptive Noise Inversion (ANI), defined as:
\begin{equation}\label{eq:8}
  \begin{aligned}
 \hat{x}_0&={\rm Decode}\left(z_0+f(t)\ast(\epsilon-\epsilon_{\theta}(z_t,t,C))\right), \\
 f(t)&=
    \begin{cases}
 \sqrt{\frac{1-\bar{\alpha}_t}{\bar{\alpha}_t}} & \text{if } t \leq R_{tmax}, \\
  
 f(R_{tmax}) & \text{if } t > R_{tmax},
    \end{cases}
  \end{aligned}
\end{equation}
where $R_{tmax}$ is the predefined constant. Building on Eq. (\ref{eq:6}), ANI directly truncates the model's predictions based on the timestep $t$.
When $t$ exceeds $R_{tmax}$,  $f(t)$ is set to $f(R_{tmax})$ to prevent the reconstructed $\hat{x}_0$ from being noisy.

\section{Additional Experimental Results}\label{sec:results}

\subsection{More Quantitative Comparisons}\label{sec:sup_compare}
As shown in the Tab. \ref{tab:sup_compare}, we provide comparisons with more methods (\eg, GAN-based, animation-based, and avatar-based). Benefiting from RGB-level expression input and simultaneous processing with identity information, our method significantly surpasses other methods.

\subsection{Combined with Text Prompts}
As depicted in Fig. \ref{fig:prompt}, we show how to combine image control and text control to generate images. Actually, prompts can be customized to meet diverse requirements while expressions are stably governed by the expression template. Moreover, an appropriate prompt will further enhance the vividness of expressions (last column).

\subsection{More Visualization Results}
\begin{figure*}[!htbp]
  \centering
  \includegraphics[width=1\textwidth]{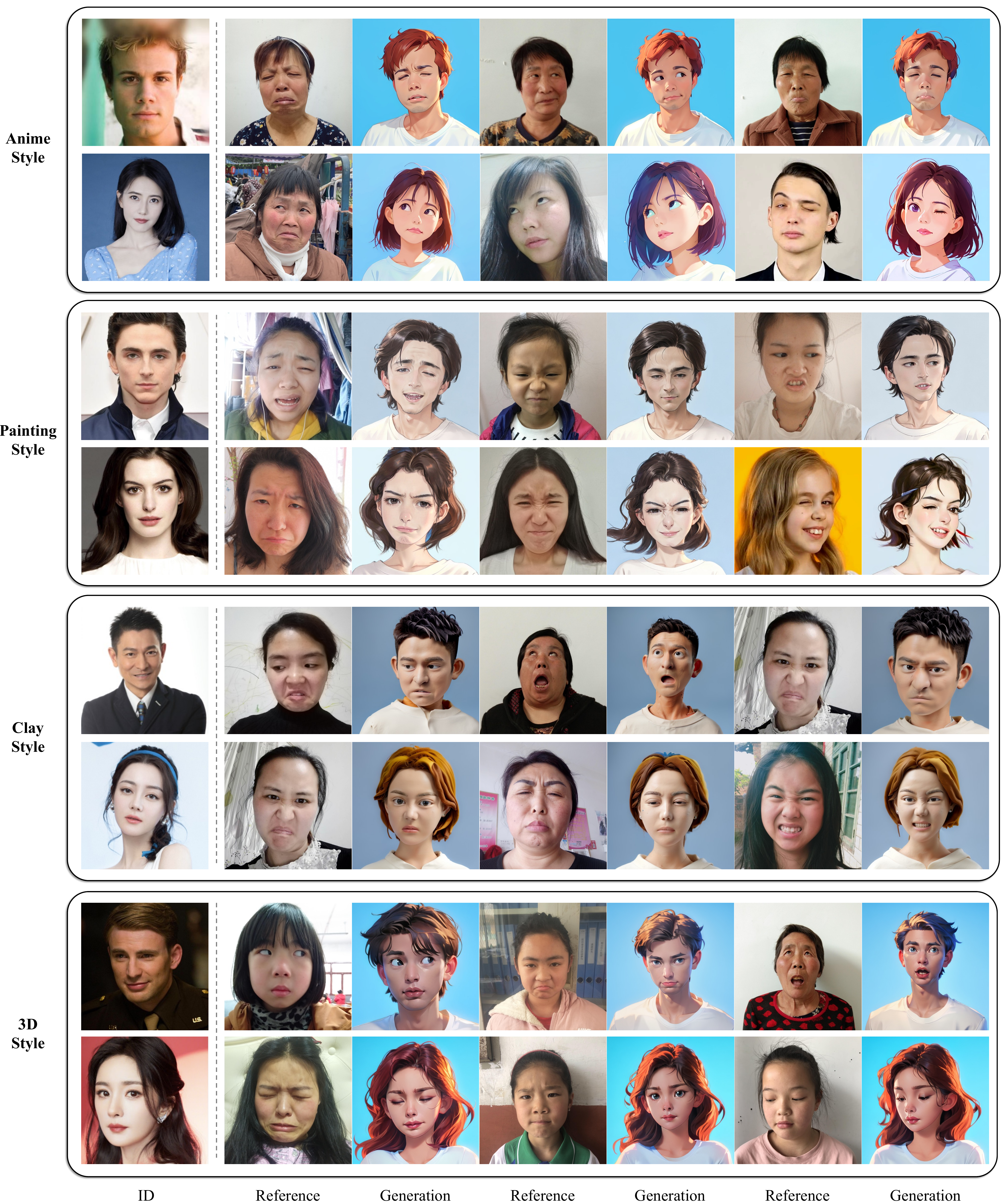}
  \caption{\textbf{More qualitative results}. For the given person (leftmost column), our method generates the corresponding image based on the various expression references, evaluated on \textbf{SD1.5} \cite{rombach2022sd} framework.
 } 
  \label{fig:res}
\end{figure*}

\begin{figure*}[!htbp]
  \centering
  \includegraphics[width=0.99\textwidth]{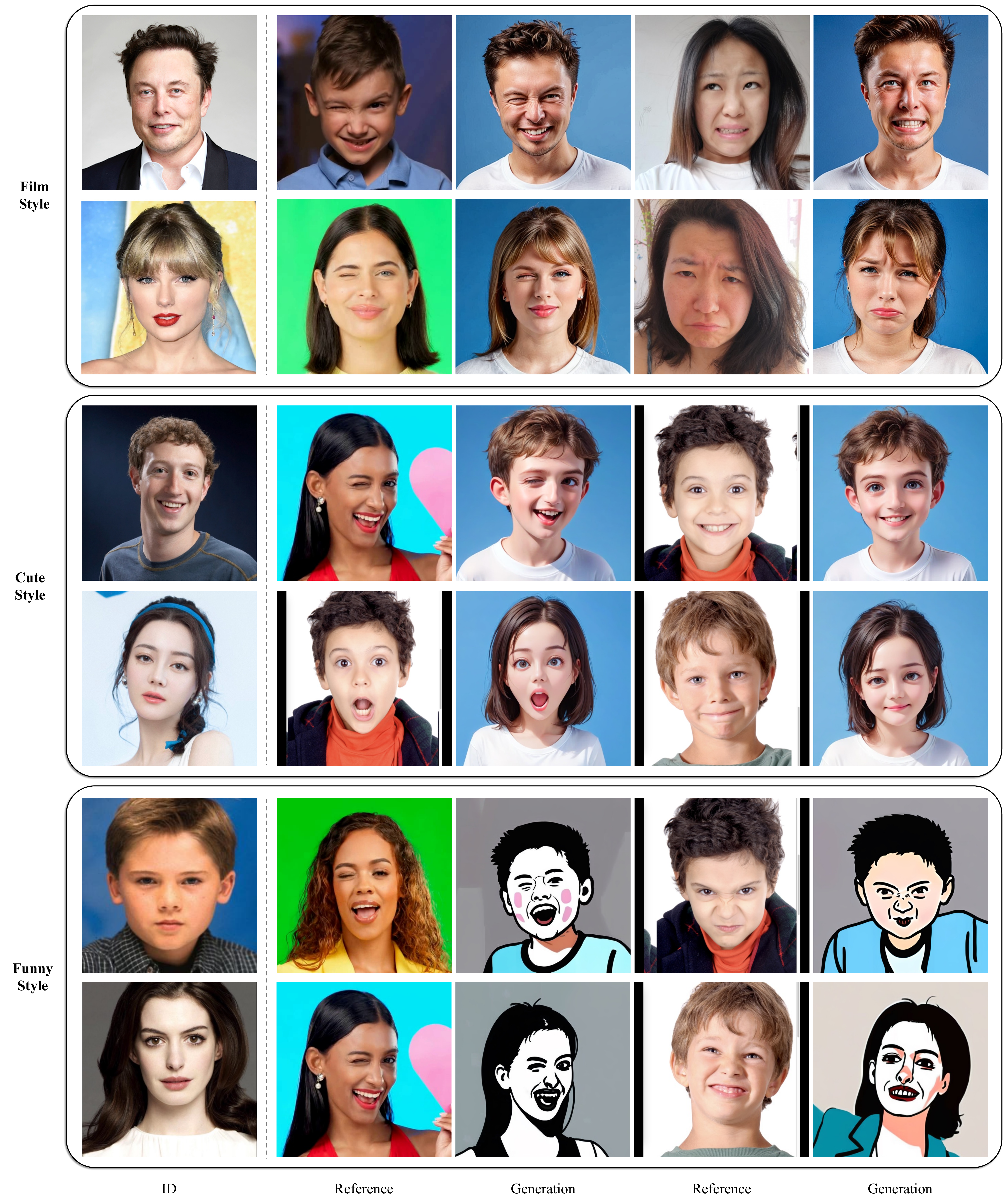}
  \caption{\textbf{More qualitative result}. For the given person (leftmost column), our method generates the corresponding image based on the various expression references, evaluated on \textbf{SDXL} \cite{podell2023sdxl} framework.
 } 
  \label{fig:res_sdxl}
\end{figure*}

In Fig. 5 of the main body, we only display a few generated images due to the page limit. In this subsection, we present additional results generated by our method on \textbf{SD1.5} \cite{rombach2022sd} and \textbf{SDXL} \cite{podell2023sdxl} framework. As shown in Fig. \ref{fig:res} and \ref{fig:res_sdxl}, our method accurately transfers the reference expression to the generated image while maintaining high identity fidelity.
\subsection{Data Visualization}
\begin{figure*}[!htbp]
  \centering
  \includegraphics[width=1\textwidth]{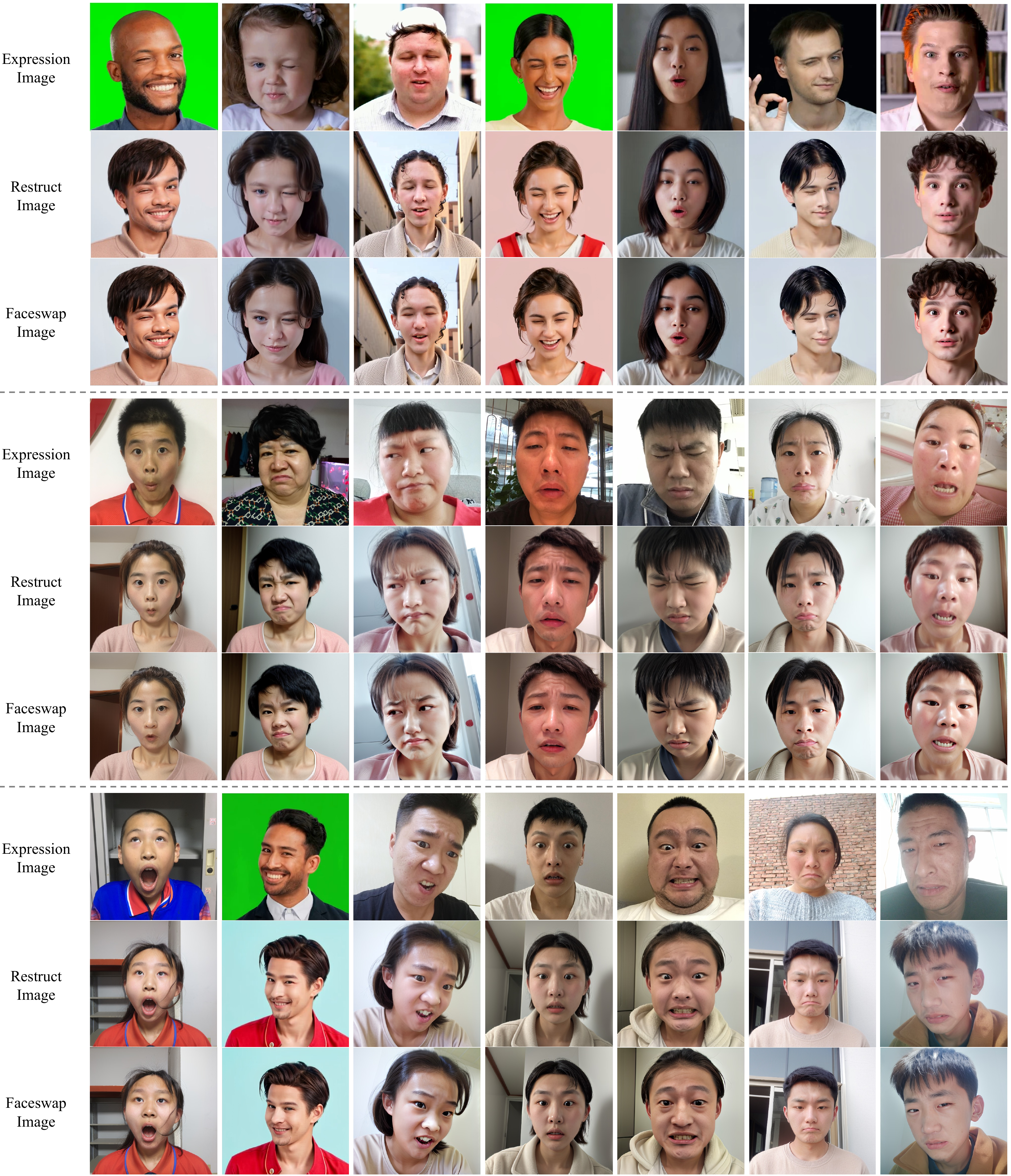}
  \caption{\textbf{More examples of IDI and CIEP100k}. ID-irrelevant Data Iteration (IDI) is introduced to transform expression reference images into the restruct and faceswap images, thus synthesizing high-quality data pairs with differing identities and consistent expressions.
 } 
  \label{fig:IDI}
\end{figure*}

In Fig. 4 of the main body, we show examples illustrating the effect of IDI in modifying the identities of individuals while maintaining facial expressions. As a supplement, we provide more visualization results as depicted in Fig. \ref{fig:IDI}, clearly demonstrating the effectiveness of our method and data quality of CIEP100k.

\end{document}